\documentclass{article}

\usepackage[final]{neurips_2024}

\usepackage{rotating}
\usepackage{lscape}
\usepackage[utf8]{inputenc} 
\usepackage[T1]{fontenc}    
\usepackage{hyperref}       
\usepackage{url}            
\usepackage{graphicx}
\usepackage{makecell}
\usepackage{array}
\usepackage{tabularx}
\usepackage[linesnumbered, ruled, vlined]{algorithm2e}
\usepackage{subcaption}
\usepackage{booktabs}       
\usepackage{amsfonts}       
\usepackage{nicefrac}       
\usepackage{microtype}      
\usepackage{multirow}
\usepackage{amsmath}
\usepackage{wrapfig}
\usepackage[table,xcdraw]{xcolor}
\usepackage{enumitem}
\usepackage{colortbl}      

\usepackage{siunitx}       
\usepackage[utf8]{inputenc}
\usepackage[T1]{fontenc}
\definecolor{lightgray}{gray}{0.9}
\definecolor{red}{rgb}{1,0,0}
\definecolor{pink}{rgb}{1.0, 0.0, 0.5} 

\def\ourmodel{MaskFactory}

\title{Mask Factory: Towards High-quality Synthetic Data Generation for Dichotomous Image Segmentation}

\author{%
  Haotian Qian$^{1}$\thanks{Equal contribution. Deng-Ping Fan served as the project leader for this work.}\quad 
  YD Chen$^{1}$\footnotemark[1]\quad 
  Shengtao Lou$^1$\quad 
  Fahad Shahbaz Khan$^{3,4}$ \\
  \textbf{Xiaogang Jin}$^1$\thanks{Corresponding author. Contact the author at \href{mailto:jin@cad.zju.edu.cn}{jin@cad.zju.edu.cn}}\quad 
  \textbf{Deng-Ping Fan}$^{2,3}$ 
  \\
  $^1$State Key Lab of CAD\&CG, Zhejiang University \quad $^2$VCIP\&CS, Nankai University\\ 
  $^3$MBZUAI \quad 
  $^4$Linköping University
  \\
}

\begin{document}

\maketitle

\begin{abstract}
Dichotomous Image Segmentation (DIS) tasks require highly precise annotations, and traditional dataset creation methods are labor intensive, costly, and require extensive domain expertise. 
Although using synthetic data for DIS is a promising solution to these challenges, current generative models and techniques struggle with the issues of scene deviations, noise-induced errors, and limited training sample variability. 
To address these issues, we introduce a novel approach, \textbf{\ourmodel{}}, which provides a scalable solution for generating diverse and precise datasets, markedly reducing preparation time and costs. 
We first introduce a general mask editing method that combines rigid and non-rigid editing techniques to generate high-quality synthetic masks.
Specially, rigid editing leverages geometric priors from diffusion models to achieve precise viewpoint transformations under zero-shot conditions, while non-rigid editing employs adversarial training and self-attention mechanisms for complex, topologically consistent modifications. 
Then, we generate pairs of high-resolution image and accurate segmentation mask using a multi-conditional control generation method.
Finally, our experiments on the widely-used DIS5K dataset benchmark demonstrate superior performance in quality and efficiency compared to existing methods. 
The code is available at \url{https://qian-hao-tian.github.io/MaskFactory/}.
\end{abstract}

\section{Introduction}
Dichotomous image segmentation (DIS) aims to accurately segment objects from natural images \cite{zheng2024birefnet,qin2022highly}, a critical task in various computer vision applications, including medical imaging \cite{liu2023instructive,chen2024bimcv,chen2024learning}, autonomous driving \cite{xu2020squeezesegv3} and connectomics research \cite{liu2024cross,chen2023self}. However, traditional methods for collecting datasets for DIS tasks are labor-intensive, costly, and require extensive domain expertise.
Recently, synthetic data has emerged as a promising solution for generating diverse and precise datasets at scale, offering a scalable and cost-effective means for model training. 
Given the importance of high-quality synthetic data for training models, developing methods to generate such datasets for DIS tasks is crucial. 
Although generative models have been employed to assist in synthetic dataset production, they face significant limitations in terms of controllability, precision, and diversity. 

Recent synthetic methods \cite{wu2024datasetdm,nguyen2024dataset,chen2023generative} utilize diffusion models \cite{rombach2022high} to synthesize image-mask pairs from textual cues and attention maps. 
Despite progress, they still encounter three major challenges:
(1) \textit{Controllability:} Text-guided generation scheme may deviate from real-world scenes, especially with fine-grained labels, leading to uncontrollable generated images.
(2) \textit{Precision:} 
The use of attention maps can introduce noise, which degrades mask fidelity and poses challenges for applications requiring precise pixel-level segmentation, such as DIS \cite{qin2022highly}.
3. \textit{Diversity:} 
Previous works \cite{toker2024satsynth,yang2024freemask,chen2024tokenunify} are relatively uniform, with limited ability to generate diverse training samples, which makes them unsuitable for providing sufficient variability in training data.

Generating diverse human images often poses challenge due to the complexity of prompts, which limits their applicability in tasks requiring extensive variability.
It's worthy noting that the expansion of datasets with fine granularity, such as DIS datasets, currently lacks an effective solution.
In fact, we observe that existing DIS datasets predominantly consist of single-source orthophotos featuring limited shape diversity among similar objects. The generation of more diverse DIS images that accurately reflect real-world distributions remains an unresolved challenge.

\begin{figure}[t]
    \centering
    \makebox[\textwidth][c]{\includegraphics[width=1.0\textwidth]{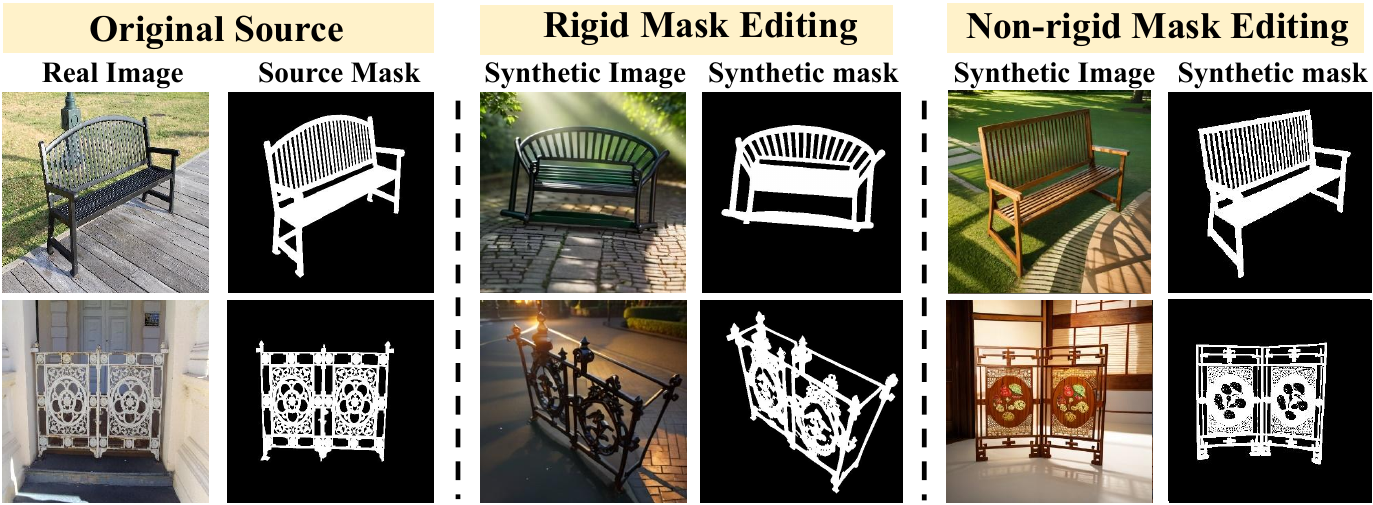}}
    \caption{Shows the edited masks from the first stage and the corresponding images generated in the second stage. In the examples, we transformed the viewpoint of park benches and tables from a frontal view to a top-down view and edited their shapes, changing park benches from curved to square edges and tables from square to circular shapes.}
    \label{fig:maskedit}
\end{figure}

To address these challenges, we propose MaskFactory, a novel two-stage method that efficiently generates high-quality synthetic datasets for DIS tasks.
Based on the geometric characteristics of objects, our approach simultaneously considers both rigid and non-rigid transformations of target objects. This means that when generating synthetic datasets, we take into account not only changes in viewpoints (rigid transformations) but also deformations (non-rigid transformations), as illustrated in Figure \ref{fig:maskedit}.
In the first stage, rigid transformations are driven by geometric priors learned from large-scale diffusion models, enabling precise viewpoint changes and simulating diverse variations in observation angles and scales. Non-rigid transformations leverage prompts provided by large language models to accurately alter the shape of target objects via attention-based editing, ensuring topological consistency before and after editing through topology-preserving adversarial training. Consequently, even for masks with complex geometries, high-quality and diverse synthetic masks can be generated for DIS tasks.
In the second stage, we utilize multiple control inputs, including masks, canny edges, and prompts representing class information, to guide the generation process. These inputs are fed into diffusion models to produce high-resolution images that match the prepared segmentation masks. This approach ensures high consistency between images and masks while enhancing the realism and diversity of the dataset.

Our method significantly enhances the realism and diversity of generated datasets. We validate the effectiveness of MaskFactory on the DIS5K dataset \cite{qin2022highly}, specifically designed to evaluate DIS performance. Experimental results show that our approach outperforms existing dataset synthesis methods in terms of structural preservation and error metrics, achieving an average performance gain of 8.8\%. These findings highlight the potential of MaskFactory to provide the data diversity and annotation precision required for DIS tasks while significantly reducing the time and costs associated with dataset preparation.


In summary, the {contributions} of our work are listed as follows:
\vspace{-0.5em}
\begin{itemize}[leftmargin=1em]
\setlength{\itemsep}{0em}
\item  We introduce \ourmodel{}, a novel approach that generate high-quality datasets for DIS task in terms of quality, precision, and efficiency.
\item We propose a two-step method that synthesizes high-quality and diverse object masks via masking editing and generates corresponding high-resolution images using a multi-conditional control generation method.
\item Experimental results on the DIS5K dataset demonstrate the superior performance of \ourmodel{}, compared to existing dataset synthesizing methods. 
\end{itemize}

\begin{figure*}[t]
\begin{center}
\includegraphics[width=\linewidth]{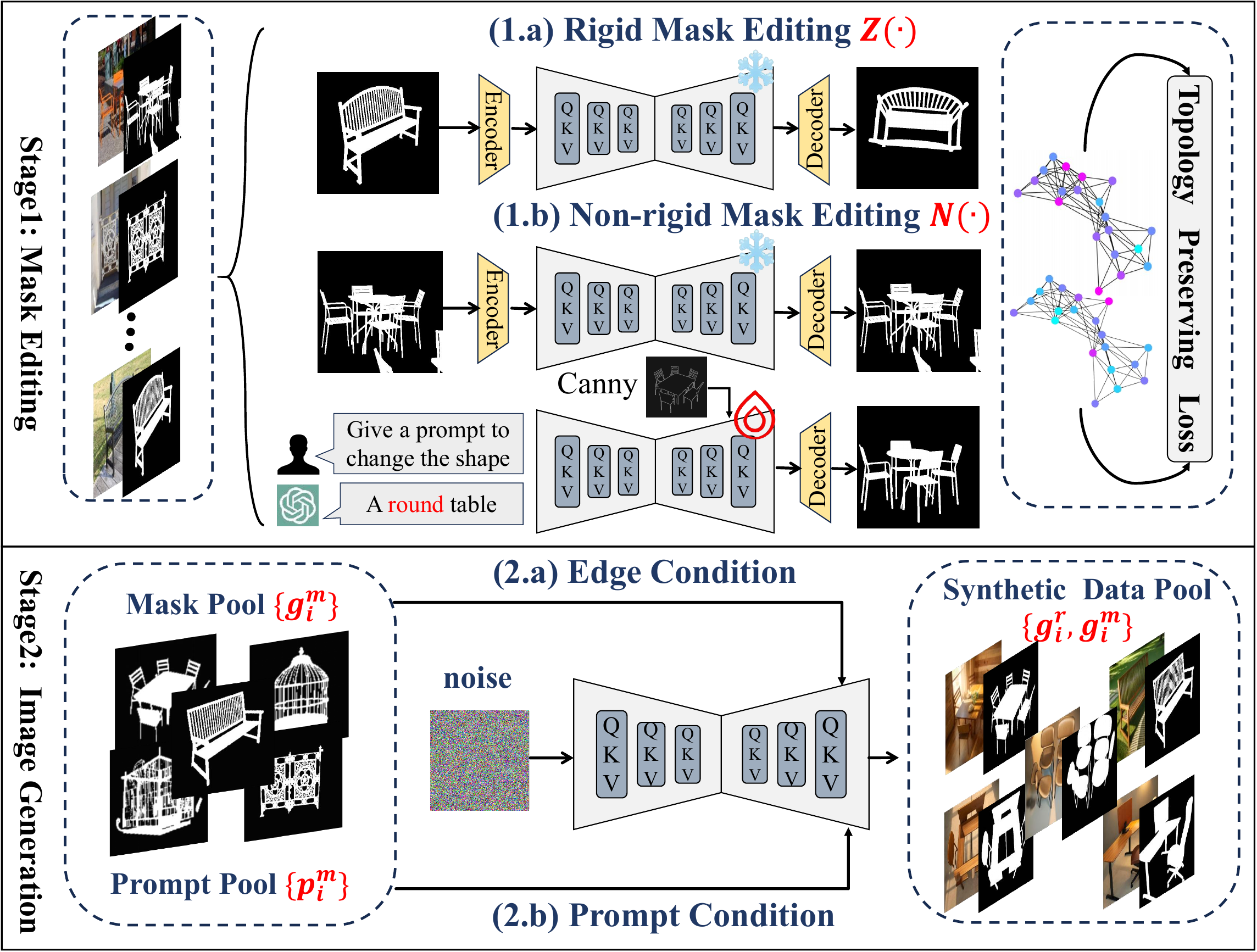}
\end{center}
\vspace{-0.5em}
\caption{Workflow of MaskFactory. In the first stage, we generate new masks by applying rigid and non-rigid editing to the existing ground truth masks. In the second stage, we use the generated masks and their corresponding extracted Canny edges as conditions, along with a prompt representing the category, to generate RGB images. This process forms paired data for our generative model.} 
\label{fig:model}
\vspace{-.15in}
\end{figure*}


\section{Related work}
\paragraph{Synthetic data.}
Synthetic data has garnered significant attention in various machine learning and computer vision tasks, such as natural language processing \cite{wang2018synthetic}, object detection \cite{lin2023explore}, and image segmentation \cite{nguyen2024dataset}. For instance, Lin \emph{et al.} \cite{lin2023explore} demonstrated that synthetic data can enhance performance in object detection, especially in scenarios with limited access to real-world data.
Generative adversarial networks (GANs) and variational autoencoders (VAEs) are two popular deep learning-based methods for generating synthetic data \cite{figueira2022survey}. Recent studies have focused on improving the quality and diversity of synthetic data \cite{kishore2021synthetic} and exploring its use in few-shot learning \cite{nguyen2024expt}. Additionally, synthetic data has been shown to enhance the interpretability and explainability of machine learning models \cite{liu2021synthetic}.
Benefiting from recent advancements in diffusion models \cite{rombach2022high,podell2023sdxl}, methods such as DatasetDM \cite{wu2024datasetdm} and Dataset Diffusion \cite{nguyen2024dataset} can generate high-quality synthetic data for various computer vision tasks. However, these methods may introduce additional errors due to the inclusion of pseudo-labels. These errors arise because pseudo-labels can be noisy and inaccurate, leading to suboptimal training data.
Although some research \cite{chen2024towards} employs controlnet-like control schemes \cite{zhang2023adding} to mitigate these issues, the generated synthetic data may still suffer from noise and errors, making it unsuitable for the DIS task. Therefore, in this paper, we focus on generating high-quality synthetic data for the DIS task, where the generated image-mask pairs need to be highly precise and accurate.

\paragraph{Dichotomous image segmentation.}
DIS has seen substantial progress with the advent of high-resolution imaging technologies. Representative approaches include the intermediate supervision strategy in IS-Net \cite{qin2022highly}, the frequency prior method \cite{zheng2024birefnet}, and the unite-divide-unite strategy by \cite{pei2023unite}. However, these methods have enhanced segmentation accuracy but often struggle to capture extremely fine details. Thus, recent progressive refinement strategies, \emph{e.g.,} BASNet \cite{qin2019basnet}, emphasize the importance of auxiliary information such as gradient maps and multi-scale inputs. These methods propose to use gradient features and ground truth supervision to enhance the learning of weak features in complex regions. Then, BiRefNet \cite{zheng2024birefnet} maintains high-resolution inputs and employs a bilateral reference framework to better capture intricate details. The latest works, such as the multi-view aggregation network by Yu et al. \cite{yu2024multi} and the interactive segmentation approach by Liu et al. \cite{liu2024rethinking}, further advance the field by integrating diverse prompts and enhancing feature representation for high-quality segmentation.

\section{Method}

\subsection{Overview of MaskFactory}

Current image segmentation methods are significantly constrained by their dependence on limited manually annotated data, which hampers both performance and generalization. 
To mitigate this, pseudo-label generation is often employed to augment training datasets. 
However, in the context of DIS, these methods frequently introduce artifacts that degrade segmentation quality. 
Additionally, existing image editing techniques often fail to preserve the topological structure of binary masks, resulting in discontinuities and overlaps within target regions.

To address these challenges, we introduce the MaskFactory framework, designed to generate a large number of high-quality synthetic image-mask pairs $\mathcal{G}=\{(g_{i}^{r},g_{i}^{m})\}_{i=1}^{M}$ from an original dataset $\mathcal{D}=\{(I_{i}^{r}, I_{i}^{m})\}_{i=1}^{N}$. This approach aims to enhance the performance of DIS models. As illustrated in Figure \ref{fig:model}, our framework comprises two main stages: mask editing and image generation.

In the mask editing stage, source masks from the original dataset are transformed using both rigid and non-rigid editing methods, resulting in a set of high-precision synthetic masks $\mathcal{G}^{m}=\{g_{i}^{m}\}_{i=1}^{M}$. Rigid mask editing generates synthetic masks from various perspectives, while non-rigid mask editing employs a topology-preserving adversarial training mechanism to edit masks according to semantic prompts while retaining the structural integrity of the source masks.

In the image generation stage, a multi-conditional control generation method is utilized to produce realistic RGB images $\mathcal{G}^{r}=\{g_{i}^{r}\}_{i=1}^{M}$ that correspond precisely to the synthetic masks, using the latter as conditioning constraints.

\subsection{Mask Editing Stage}

\subsubsection{Rigid Mask Editing}

Rigid mask editing aims to preserve detailed information from the source masks through rigid transformations. We leverage the Zero123 \cite{liu2023zero} method, which employs a viewpoint-conditioned diffusion model $\psi_{\theta}$ to manipulate masks' perspectives. Given the relative camera rotation and translation $\mathbf{T}_i$ for the desired viewpoint, $\psi_{\theta}$ synthesizes a new mask $g_{i}^{m}$ based on the source mask $I_{i}^{m}$, such that $g_i^m=\psi_{\theta}(I_i^{m'},\mathbf{T}_i)$, where $I_i^{m'}$ is the inverted image of the source mask $I_{i}^{m}$ to ensure the main component is non-zero.

\subsubsection{Non-Rigid Mask Editing}

Non-rigid mask editing, inspired by MasaCtrl \cite{cao2023masactrl}, is a critical component of MaskFactory. Unlike MasaCtrl, which directly manipulates the source mask $I_i^m$ using a textual prompt $P_i$ to generate a synthetic mask $g_{i}^{m}$, we introduce a topology-preserving adversarial training mechanism to mitigate artifacts and structural degradation in binary mask editing. The textual prompt $P_i$ is derived from a pool of prompts \{$p^m_i$\} that are generated using GPT-4 based on the original images. These prompts provide detailed descriptions that guide the mask editing process. This module consists of a generator $G_{\theta}$ and a discriminator $D_{\phi}$. The generator transforms noise $\mathbf{z}$ into a synthetic mask $g_i^m$ under the guidance of a textual prompt $P_i$ and the source mask $I_i^m$. A mutual attention mechanism aligns the query features $\mathbf{Q}_t$ of $g_i^m$ with the key and value features $\mathbf{K}_s, \mathbf{V}_s$ of $I_i^m$, ensuring consistency during editing. To avoid foreground-background confusion, a mask $\mathbf{M}$ is extracted from the cross-attention maps to guide the model's focus.

\paragraph{Topology-Preserving Adversarial Training.}

To maintain the structural information of the source mask, we first extract an edge map $E_s = \mathcal{E}(I_i^m)$ using an edge detection operator $\mathcal{E}$, obtaining key points $\mathcal{V} = \{v_j\}_{j=1}^{N_v}$. We then construct a graph $\mathcal{T} = (\mathcal{V}, \mathcal{E}_s)$ based on these key points. The discriminator $D_{\phi}$ performs adversarial training on the structural graphs $\mathcal{T}_{g}$ and $\mathcal{T}_{s}$ of the synthetic mask $g_i^m$ and the source mask, respectively, ensuring topological consistency.

The training objective of the discriminator $D_{\phi}$ is to maximize:
\begin{equation}
\max_{\phi} \mathbb{E}_{\mathcal{T}_{s} \sim p_{\text{data}}(\mathcal{T}_{s})}[\log D_{\phi}(\mathcal{T}_{s})] + \mathbb{E}_{\mathcal{T}_{g} \sim p_{\text{gen}}(\mathcal{T}_{g})}[\log (1 - D_{\phi}(\mathcal{T}_{g}))],
\end{equation}
where $ p_{\text{data}}(\mathcal{T}_{s}) $ and $ p_{\text{gen}}(\mathcal{T}_{g}) $ represent the distributions of the structural graphs of the source masks and the synthetic masks, respectively. Conversely, the training objective of the generator $ G_{\theta} $ is to minimize the discriminative power of the discriminator:
\begin{equation}
\min_{\theta} \mathbb{E}_{\mathcal{T}_{g} \sim p_{\text{gen}}(\mathcal{T}_{g})}[\log (1 - D_{\phi}(\mathcal{T}_{g}))].
\end{equation}

Through topology-preserving adversarial training, the non-rigid editing module effectively retains the structural information from the source masks during the editing process, generating high-quality, artifact-free synthetic masks.

The overall loss function $\mathcal{L}_{\text{total}}$ for non-rigid mask editing encompasses the adversarial loss $\mathcal{L}_{\text{GAN}}$, the content loss $\mathcal{L}_{\text{content}}$, and the structure preservation loss $\mathcal{L}_{\text{structure}}$:
\begin{equation}
\mathcal{L}_{\text{total}} = \mathcal{L}_{\text{GAN}} + \lambda_1 \mathcal{L}_{\text{content}} + \lambda_2 \mathcal{L}_{\text{structure}},
\label{equ:lambda}
\end{equation}
where $\lambda_1$ and $\lambda_2$ are balancing factors.

The adversarial loss $\mathcal{L}_{\text{GAN}}$ is defined as:
\begin{equation}
\mathcal{L}_{\text{GAN}} = \mathbb{E}_{\mathcal{T}_{s} \sim p_{\text{data}}(\mathcal{T}_{s})}[\log D_{\phi}(\mathcal{T}_{s})] + \mathbb{E}_{\mathcal{T}_{g} \sim p_{\text{gen}}(\mathcal{T}_{g})}[\log (1 - D_{\phi}(\mathcal{T}_{g}))].
\end{equation}

The content loss $\mathcal{L}_{\text{content}}$ measures the semantic consistency between the synthetic mask and the textual prompt:
\begin{equation}
\mathcal{L}_{\text{content}} = \lVert g_i^m - I_i^m \rVert_1,
\end{equation}
where $ g_i^m $ is the synthetic mask and $ I_i^m $ is the source mask.

The structure preservation loss $\mathcal{L}_{\text{structure}}$ evaluates the difference between the structural graphs of the synthetic mask and the source mask:
\begin{equation}
\mathcal{L}_{\text{structure}} = \lVert \mathcal{T}_{g} - \mathcal{T}_{s} \rVert_1.
\end{equation}

These components ensure that the editing process maintains the structural and semantic consistency of the source masks.

\subsection{Image Generation Stage}

Following the mask editing stage, we obtain a synthetic mask pool $\{g_i^m\}_{i=0}^M$ comprising finely detailed synthetic masks generated through the aforementioned transformations. In the subsequent image generation stage, we accurately generate corresponding RGB images $\{g_i^r\}_{i=0}^M$ for the masks in the synthetic mask pool using a multi-condition control generation method. The primary segmentation of the RGB images aligns with the corresponding synthetic masks. Inspired by ControlNet \cite{zhang2023adding}, we introduce a multi-condition control generation method to achieve precise RGB image generation. This method simultaneously injects the segmentation condition $c_i^s$ and the canny condition $c_i^y$ to steer the denoising process of the random Gaussian noise, ensuring the generated RGB images $\{g_i^r\}_{i=0}^M$ correspond accurately to the synthetic masks $\{g_i^m\}_{i=0}^M$.

Since the synthetic mask itself serves as the segmentation condition $c_i^s$, when a synthetic mask $g_i^{m}$ is provided, we only need to extract the canny condition $c_i^y$ using the canny operator:
\begin{equation}
c_i^s=g_i^m, \quad c_i^y=Canny(g_i^m)
\end{equation}
After obtaining the canny and segmentation conditions, we input them into block $B_\theta$, which consists of a set of neural layers. Finally, these conditions are injected into the pre-trained diffusion model $M_\theta$, controlling the noise $z$ denoising process to generate the corresponding RGB image. We refer readers to \cite{zhang2023adding} for more details.
\begin{equation}
g_i^r=M_\theta(z, B_\theta(c_i^s), B_\theta(c_i^y))
\end{equation}


\begin{figure}[t]
    \centering
    \includegraphics[width=1.0\linewidth]{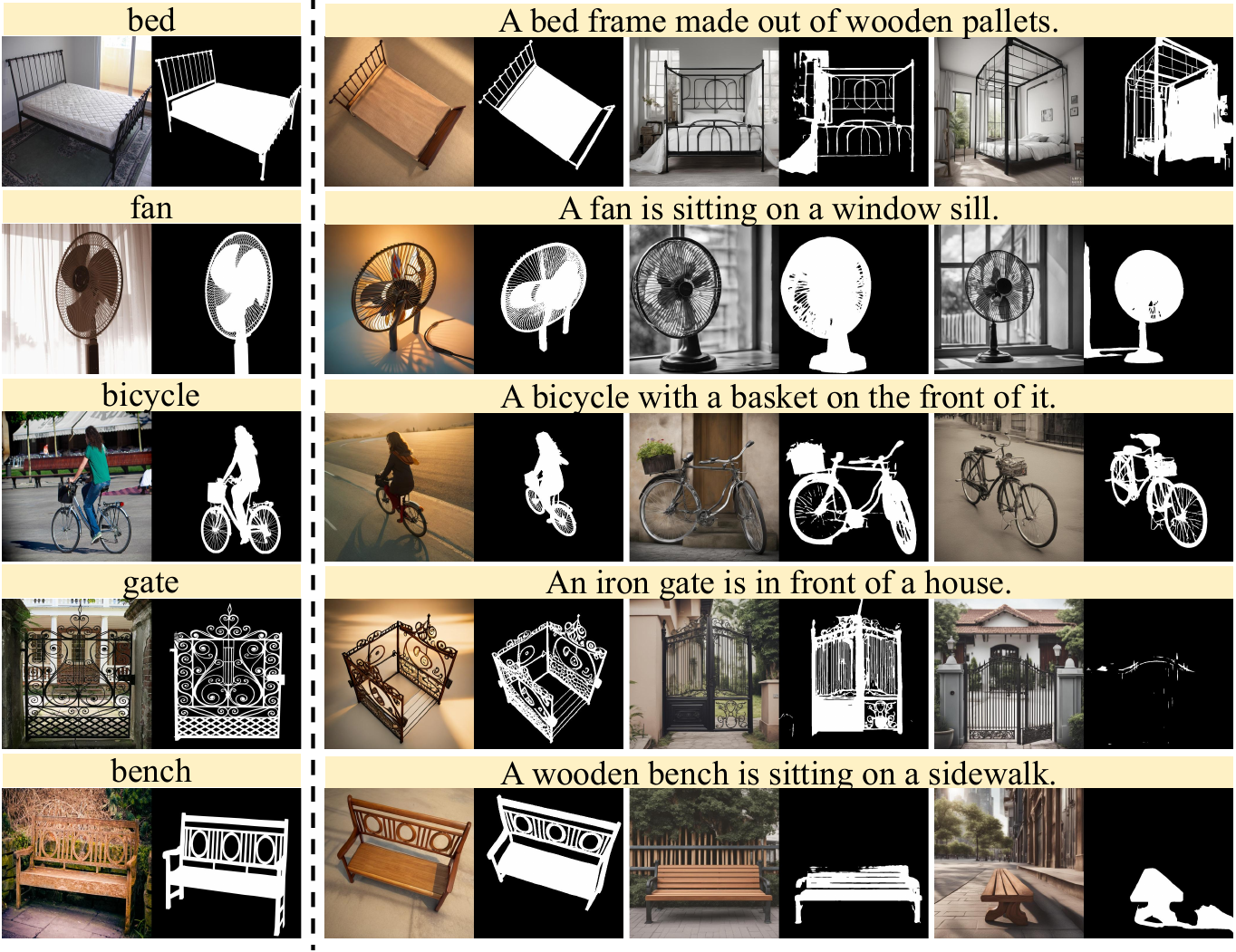}
        \put(-0.96\linewidth, -0.4em){(a) Raw Image}
    \put(-0.71\linewidth, -0.4em){(b) MaskFactory}
    \put(-0.47\linewidth, -0.4em){(c) DatasetDM \cite{wu2024datasetdm}}
    \put(-0.25\linewidth, -0.4em){(d) DatasetDiffusion\cite{nguyen2024dataset}}
    \caption{compared with baseline methods}
    \vspace{-0.3cm}
    \label{fig:baseline}
\end{figure}
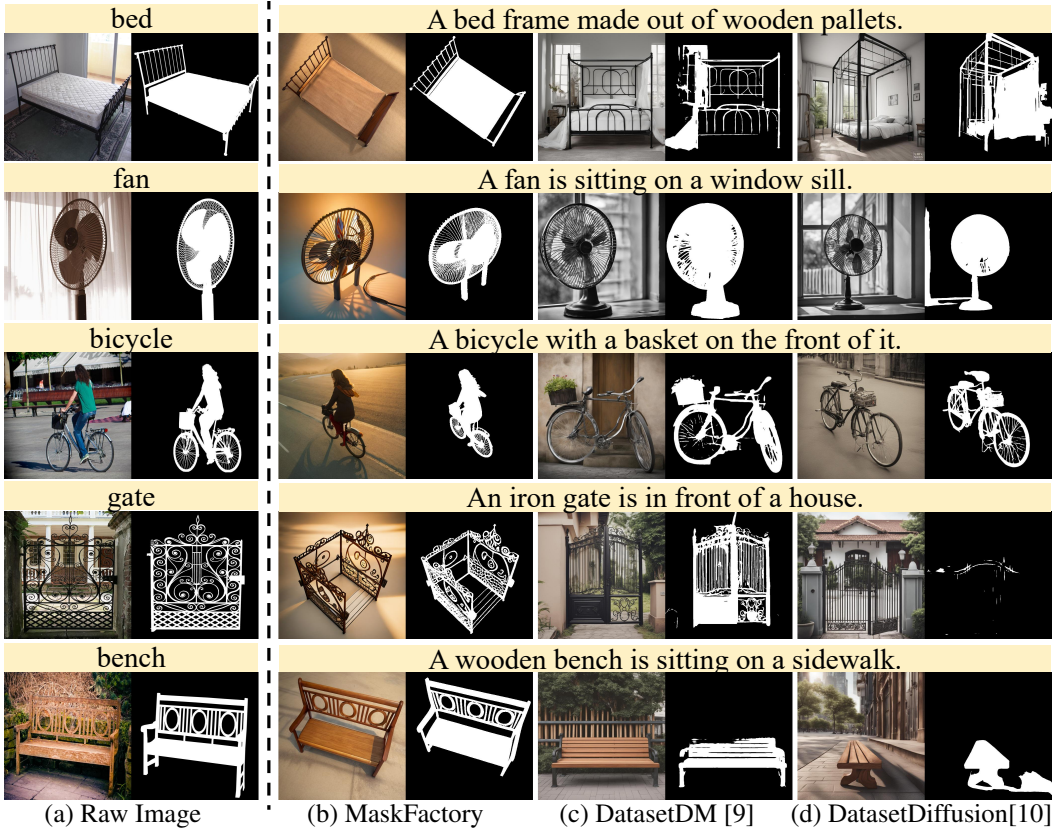


\section{Experiment and Results}
\vspace{-0.3cm}
\subsection{Dataset \& Metrics}
We conduct our experiments on the DIS5K dataset, which comprises 5,479 high-resolution images featuring camouflaged, salient, or meticulous objects in various backgrounds. The DIS5K dataset is divided into three subsets: DIS-TR (3,000 images) for training, DIS-VD (470 images) for validation, and DIS-TE (2,000 images) for testing. For data augmentation, we utilize the mask portion of the training subset (DIS-TR).

To evaluate our models, we employ a diverse set of metrics to ensure comprehensive performance assessment. These metrics include {max F\(_1\)}\cite{achanta2009frequency}, which balances precision and recall, providing a harmonic mean that is indicative of overall accuracy; {F\(_\beta^\omega\)}\cite{margolin2014evaluate}, a weighted F-measure that compensates for class imbalances, with values ranging from 0 to 1, where higher values denote superior performance; {$M$} (Mean Absolute Error)\cite{perazzi2012saliency}, which calculates the average absolute difference between the predicted and ground truth masks, with lower values signifying better accuracy; {S\(_\alpha\)}\cite{fan2017structure}, a structural similarity measure that evaluates the preservation of significant structures within the image, with values closer to 1 indicating better performance; and {E\(_M^\phi\)}\cite{fan2018enhanced}, an enhanced measure that considers both pixel-level and image-level information for a more holistic evaluation, where higher values represent better performance. Collectively, these metrics provide a robust framework for assessing the effectiveness and reliability of our segmentation models.

\begin{table}[t]
\fontsize{8.5}{8}\selectfont
\centering
\renewcommand\tabcolsep{2.5pt}
\caption{Comparison of experimental results with different generation methods, illustrating the impact of using various numbers of generated images on segmentation task performance.}
\label{tab:main1}
\begin{tabularx}{\linewidth}{rrc|*{4}{X}|*{4}{X}|*{4}{X}}
\toprule[1.2pt]
\multicolumn{2}{r}{\multirow{2}{*}{\textbf{Dataset}}}                                                      & \multirow{2}{*}{\textbf{IS-Net}} & \multicolumn{4}{c|}{\textbf{DatasetDM} \cite{wu2024datasetdm}}    & \multicolumn{4}{c|}{\textbf{Dataset Diffusion} \cite{nguyen2024dataset}} & \multicolumn{4}{c}{\textbf{MaskFactory}}                                              \\
\multicolumn{2}{r}{}                                                                              &                         & 2500 & 5000 & 7500 & 10000 & 2500 & 5000 & 7500 & 10000 & 2500 & 5000 & 7500 & 10000 \\
\midrule
\multirow{5}{*}{\rotatebox{90}{DIS-VD}} & $max F_1 \uparrow$ & 0.791 & 0.792 & 0.791 & 0.788 & 0.785 & 0.793 & 0.780 & 0.771 & 0.767 & 0.831 & 0.833 & 0.834 & \textbf{0.835} \\
& $F_{\beta}^{\omega} \uparrow$ & 0.717 & 0.720 & 0.719 & 0.712 & 0.710 & 0.726 & 0.726 & 0.716 & 0.710 & 0.725 & 0.754 & 0.757 & \textbf{0.759} \\
& $M \downarrow$ & 0.074 & 0.076 & 0.076 & 0.075 & 0.077 & 0.074 & 0.077 & 0.076 & 0.077 & 0.073 & 0.073 & \textbf{0.071} & 0.072 \\
& $S_{\alpha} \uparrow$ & 0.813 & 0.814 & 0.810 & 0.807 & 0.805 & 0.826 & 0.821 & 0.804 & 0.790 & 0.832 & 0.855 & 0.860 & \textbf{0.866} \\
& $E_M^{\phi} \uparrow$ & 0.856 & 0.869 & 0.864 & 0.864 & 0.860 & 0.868 & 0.859 & 0.852 & 0.838 & 0.880 & 0.911 & 0.914 & \textbf{0.923} \\
\midrule
\multirow{5}{*}{\rotatebox{90}{DIS-TE1}} & $max F_1 \uparrow$ & 0.740 & 0.744 & 0.744 & 0.740 & 0.736 & 0.741 & 0.727 & 0.721 & 0.719 & 0.776 & 0.777 & 0.779 & \textbf{0.784} \\
& $F_{\beta}^{\omega} \uparrow$ & 0.662 & 0.670 & 0.668 & 0.663 & 0.659 & 0.675 & 0.669 & 0.667 & 0.654 & 0.677 & 0.700 & 0.703 & \textbf{0.705} \\
& $M \downarrow$ & 0.074 & 0.075 & 0.077 & 0.075 & 0.076 & 0.076 & 0.074 & 0.077 & 0.078 & 0.073 & \textbf{0.071} & \textbf{0.071} & 0.073 \\
& $S_{\alpha} \uparrow$ & 0.787 & 0.791 & 0.786 & 0.781 & 0.778 & 0.790 & 0.776 & 0.760 & 0.754 & 0.803 & 0.822 & 0.826 & \textbf{0.829} \\
& $E_M^{\phi}\uparrow$ & 0.820 & 0.825 & 0.819 & 0.813 & 0.810 & 0.827 & 0.816 & 0.802 & 0.798 & 0.853 & 0.866 & 0.868 & \textbf{0.875} \\
\midrule
\multirow{5}{*}{\rotatebox{90}{DIS-TE2}} & $max F_1 \uparrow$ & 0.799 & 0.810 & 0.807 & 0.801 & 0.793 & 0.801 & 0.799 & 0.782 & 0.776 & 0.808 & 0.814 & 0.819 & \textbf{0.822} \\
& $F_{\beta}^{\omega} \uparrow$ & 0.728 & 0.741 & 0.736 & 0.736 & 0.734 & 0.739 & 0.737 & 0.721 & 0.712 & 0.739 & 0.764 & 0.765 & \textbf{0.772} \\
& $M \downarrow$ & 0.070 & 0.072 & 0.070 & 0.071 & 0.071 & 0.072 & 0.071 & 0.073 & 0.071 & 0.068 & 0.067 & 0.068 & \textbf{0.067} \\
& $S_{\alpha} \uparrow$ & 0.826 & 0.833 & 0.828 & 0.823 & 0.821 & 0.833 & 0.832 & 0.811 & 0.795 & 0.849 & 0.855 & 0.861 & \textbf{0.865} \\
& $E_M^{\phi} \uparrow$ & 0.858 & 0.870 & 0.868 & 0.866 & 0.865 & 0.861 & 0.860 & 0.850 & 0.834 & 0.868 & 0.895 & 0.901 & \textbf{0.903} \\
\midrule
\multirow{5}{*}{\rotatebox{90}{DIS-TE3}} & $max F_1 \uparrow$ & 0.830 & 0.846 & 0.845 & 0.839 & 0.833 & 0.838 & 0.834 & 0.831 & 0.812 & 0.841 & 0.850 & 0.867 & \textbf{0.870} \\
& $F_{\beta}^{\omega} \uparrow$ & 0.758 & 0.770 & 0.766 & 0.763 & 0.757 & 0.769 & 0.752 & 0.733 & 0.728 & 0.761 & 0.782 & 0.783 & \textbf{0.785} \\
& $M \downarrow$ & 0.064 & 0.066 & 0.066 & 0.065 & 0.066 & 0.066 & 0.065 & 0.066 & 0.065 & \textbf{0.062} & \textbf{0.062} & \textbf{0.062} & 0.063 \\
& $S_{\alpha} \uparrow$ & 0.836 & 0.848 & 0.844 & 0.839 & 0.832 & 0.844 & 0.837 & 0.823 & 0.819 & 0.842 & 0.859 & 0.866 & \textbf{0.878} \\
& $E_M^{\phi} \uparrow$ & 0.883 & 0.894 & 0.887 & 0.883 & 0.878 & 0.900 & 0.896 & 0.878 & 0.858 & 0.911 & 0.915 & 0.918 & \textbf{0.926} \\
\midrule
\multirow{5}{*}{\rotatebox{90}{DIS-TE4}} & $max F_1 \uparrow$ & 0.827 & 0.833 & 0.826 & 0.820 & 0.816 & 0.830   & 0.818  & 0.804  & 0.801  & 0.856          & 0.876          & 0.876          & \textbf{0.879} \\
                         & $F_{\beta}^{\omega} \uparrow$ & 0.753 & 0.759 & 0.754 & 0.750 & 0.746 & 0.760   & 0.759  & 0.758  & 0.747  & 0.789          & 0.824          & 0.824          & \textbf{0.830} \\
                         & $M \downarrow$               & 0.072 & 0.074 & 0.073 & 0.073 & 0.075 & 0.074   & 0.072  & 0.076  & 0.074  & 0.071          & \textbf{0.069} & 0.071          & 0.072          \\
                         & $S_{\alpha} \uparrow$        & 0.830 & 0.846 & 0.839 & 0.838 & 0.833 & 0.841   & 0.830  & 0.822  & 0.815  & 0.842          & 0.852          & 0.860          & \textbf{0.862} \\
                         & $E_M^{\phi} \uparrow$        & 0.870 & 0.885 & 0.881 & 0.873 & 0.868 & 0.873   & 0.866  & 0.845  & 0.836  & 0.891          & 0.917          & 0.917          & \textbf{0.923} \\ \bottomrule[1.2pt]
\end{tabularx}
\end{table}

\subsection{Impletement Details}
\label{details}
Our image editing framework is implemented using PyTorch and trained on 8 NVIDIA GeForce RTX 3090 GPUs. We employ both non-rigid and rigid editing approaches to manipulate images in the DIS-TR dataset, utilizing spatial transformers and affine transformations to model non-rigid and rigid deformations, respectively. The hyperparameters used in our model are as follows: a batch size of 16, an image size of 512x512, 5 editing iterations, a learning rate of 0.001, a weight decay of 0.0001, 1000 diffusion steps, and a diffusion step size of 0.1. The hyperparameters in Equ \ref{equ:lambda} are set to $\lambda_1=0.8$ and $\lambda_2=0.5$.

We train the segmentation model using the DIS-TR subset of the DIS5K dataset, utilizing 2 NVIDIA GeForce RTX 3090 GPUs. The input size to the network is 512x512, with a learning rate of 0.0001 and a batch size of 48. The model is optimized using the Adam optimizer over a total of 800 epochs. The edited images generated by our model are combined with the original training set to form a new training set, enabling the model to learn from both original and edited images.

We evaluate our model on the DIS5K datasets, using the test sets from each dataset. For a comprehensive comparison, we also reproduce two state-of-the-art diffusion-based image generation methods, DatasetDM and Dataset Diffusion, using their default parameters. The generated images are compared with those from our approach, with the visualization of the results shown in Figure \ref{fig:baseline}.

\begin{table}[t]
 \fontsize{8.5}{9.6}\selectfont
        \centering
        \renewcommand\tabcolsep{3pt}
        \caption{Comparison of experimental results using different segmentation schemes, demonstrating performance gains with a fixed number of generated images.}
        \label{tab:main2}
 \begin{tabularx}{\linewidth}{rr|*{10}{c}}
\toprule[1.2pt]
\multicolumn{2}{r|}{}                                 & \multicolumn{2}{c}{\textbf{IS-Net \cite{qin2022highly}}}                       & \multicolumn{2}{c}{\textbf{FP-DIS \cite{zhou2023dichotomous}}}                       & \multicolumn{2}{c}{\textbf{UDUN \cite{pei2023unite}}}                         & \multicolumn{2}{c}{\textbf{BiRefNet \cite{zheng2024birefnet}}}                     & \multicolumn{2}{c}{\textbf{SAM-HQ \cite{ke2024segment}}}                       \\ 
\multicolumn{2}{r|}{\multirow{-2}{*}{\textbf{Method}}}         & w/o Ours & w Ours                & w/o Ours & w Ours                & w/o Ours & w Ours                & w/o Ours & w Ours                & w/o Ours & w Ours                \\ \midrule
                          & $max F_1 \uparrow$                 & 0.740            & \cellcolor[HTML]{EFEFEF}0.784 & 0.784            & \cellcolor[HTML]{EFEFEF}0.805 & 0.784            & \cellcolor[HTML]{EFEFEF}0.799 & 0.866            & \cellcolor[HTML]{EFEFEF}0.882 & 0.897            & \cellcolor[HTML]{EFEFEF}0.905 \\
                          & $M \downarrow$                       & 0.074            & \cellcolor[HTML]{EFEFEF}0.073 & 0.060            & \cellcolor[HTML]{EFEFEF}0.063 & 0.059            & \cellcolor[HTML]{EFEFEF}0.057 & 0.036            & \cellcolor[HTML]{EFEFEF}0.033 & 0.019            & \cellcolor[HTML]{EFEFEF}0.018 \\
                          & $S_{\alpha} \uparrow$                   & 0.787            & \cellcolor[HTML]{EFEFEF}0.829 & 0.821            & \cellcolor[HTML]{EFEFEF}0.859 & 0.817            & \cellcolor[HTML]{EFEFEF}0.830 & 0.889            & \cellcolor[HTML]{EFEFEF}0.900 & 0.907            & \cellcolor[HTML]{EFEFEF}0.911 \\
\multirow{-4}{*}{\rotatebox{90}{DIS-TE1}} & $E_M^{\phi} \uparrow$ & 0.820            & \cellcolor[HTML]{EFEFEF}0.875 & 0.855            & \cellcolor[HTML]{EFEFEF}0.885 & 0.846            & \cellcolor[HTML]{EFEFEF}0.849 & 0.915            & \cellcolor[HTML]{EFEFEF}0.916 & 0.943            & \cellcolor[HTML]{EFEFEF}0.949 \\ \midrule
                          & $max F_1 \uparrow$                 & 0.799            & \cellcolor[HTML]{EFEFEF}0.822 & 0.827            & \cellcolor[HTML]{EFEFEF}0.849 & 0.829            & \cellcolor[HTML]{EFEFEF}0.849 & 0.906            & \cellcolor[HTML]{EFEFEF}0.910 & 0.889            & \cellcolor[HTML]{EFEFEF}0.894 \\
                          & $M \downarrow$                       & 0.070            & \cellcolor[HTML]{EFEFEF}0.067 & 0.059            & \cellcolor[HTML]{EFEFEF}0.061 & 0.058            & \cellcolor[HTML]{EFEFEF}0.055 & 0.031            & \cellcolor[HTML]{EFEFEF}0.029 & 0.029            & \cellcolor[HTML]{EFEFEF}0.030 \\
                          & $S_{\alpha} \uparrow$                   & 0.823            & \cellcolor[HTML]{EFEFEF}0.865 & 0.845            & \cellcolor[HTML]{EFEFEF}0.862 & 0.843            & \cellcolor[HTML]{EFEFEF}0.866 & 0.913            & \cellcolor[HTML]{EFEFEF}0.921 & 0.883            & \cellcolor[HTML]{EFEFEF}0.889 \\
\multirow{-4}{*}{\rotatebox{90}{DIS-TE2}} & $E_M^{\phi} \uparrow$ & 0.858            & \cellcolor[HTML]{EFEFEF}0.903 & 0.889            & \cellcolor[HTML]{EFEFEF}0.893 & 0.886            & \cellcolor[HTML]{EFEFEF}0.894 & 0.947            & \cellcolor[HTML]{EFEFEF}0.957 & 0.928            & \cellcolor[HTML]{EFEFEF}0.937 \\ \midrule
                          & $max F_1 \uparrow$                 & 0.830            & \cellcolor[HTML]{EFEFEF}0.870 & 0.868            & \cellcolor[HTML]{EFEFEF}0.911 & 0.865            & \cellcolor[HTML]{EFEFEF}0.888 & 0.920            & \cellcolor[HTML]{EFEFEF}0.937 & 0.851            & \cellcolor[HTML]{EFEFEF}0.853 \\
                          & $M \downarrow$                       & 0.064            & \cellcolor[HTML]{EFEFEF}0.063 & 0.049            & \cellcolor[HTML]{EFEFEF}0.051 & 0.050            & \cellcolor[HTML]{EFEFEF}0.047 & 0.029            & \cellcolor[HTML]{EFEFEF}0.027 & 0.045            & \cellcolor[HTML]{EFEFEF}0.043 \\
                          & $S_{\alpha} \uparrow$                   & 0.836            & \cellcolor[HTML]{EFEFEF}0.878 & 0.871            & \cellcolor[HTML]{EFEFEF}0.892 & 0.865            & \cellcolor[HTML]{EFEFEF}0.885 & 0.918            & \cellcolor[HTML]{EFEFEF}0.918 & 0.851            & \cellcolor[HTML]{EFEFEF}0.854 \\
\multirow{-4}{*}{\rotatebox{90}{DIS-TE3}} & $E_M^{\phi} \uparrow$ & 0.883            & \cellcolor[HTML]{EFEFEF}0.926 & 0.903            & \cellcolor[HTML]{EFEFEF}0.924 & 0.913            & \cellcolor[HTML]{EFEFEF}0.931 & 0.951            & \cellcolor[HTML]{EFEFEF}0.957 & 0.897            & \cellcolor[HTML]{EFEFEF}0.906 \\ \midrule
                          & $max F_1 \uparrow$                 & 0.827            & \cellcolor[HTML]{EFEFEF}0.879 & 0.846            & \cellcolor[HTML]{EFEFEF}0.882 & 0.846            & \cellcolor[HTML]{EFEFEF}0.851 & 0.906            & \cellcolor[HTML]{EFEFEF}0.916 & 0.763            & \cellcolor[HTML]{EFEFEF}0.764 \\
                          & $M \downarrow$                       & 0.072            & \cellcolor[HTML]{EFEFEF}0.072 & 0.061            & \cellcolor[HTML]{EFEFEF}0.063 & 0.059            & \cellcolor[HTML]{EFEFEF}0.053 & 0.038            & \cellcolor[HTML]{EFEFEF}0.035 & 0.088            & \cellcolor[HTML]{EFEFEF}0.084 \\
                          & $S_{\alpha} \uparrow$                   & 0.830            & \cellcolor[HTML]{EFEFEF}0.862 & 0.852            & \cellcolor[HTML]{EFEFEF}0.856 & 0.849            & \cellcolor[HTML]{EFEFEF}0.873 & 0.901            & \cellcolor[HTML]{EFEFEF}0.911 & 0.799            & \cellcolor[HTML]{EFEFEF}0.806 \\
\multirow{-4}{*}{\rotatebox{90}{DIS-TE4}} & $E_M^{\phi} \uparrow$ & 0.870            & \cellcolor[HTML]{EFEFEF}0.923 & 0.891            & \cellcolor[HTML]{EFEFEF}0.935 & 0.891            & \cellcolor[HTML]{EFEFEF}0.895 & 0.933            & \cellcolor[HTML]{EFEFEF}0.941 & 0.843            & \cellcolor[HTML]{EFEFEF}0.850 \\ \midrule
                          & $max F_1 \uparrow$                 & 0.791            & \cellcolor[HTML]{EFEFEF}0.835 & 0.823            & \cellcolor[HTML]{EFEFEF}0.851 & 0.823            & \cellcolor[HTML]{EFEFEF}0.847 & 0.897            & \cellcolor[HTML]{EFEFEF}0.905 & 0.842            & \cellcolor[HTML]{EFEFEF}0.847 \\
                          & $M \downarrow$                       & 0.074            & \cellcolor[HTML]{EFEFEF}0.072 & 0.062            & \cellcolor[HTML]{EFEFEF}0.065 & 0.059            & \cellcolor[HTML]{EFEFEF}0.058 & 0.036            & \cellcolor[HTML]{EFEFEF}0.033 & 0.045            & \cellcolor[HTML]{EFEFEF}0.044 \\
                          & $S_{\alpha} \uparrow$                   & 0.813            & \cellcolor[HTML]{EFEFEF}0.866 & 0.843            & \cellcolor[HTML]{EFEFEF}0.873 & 0.838            & \cellcolor[HTML]{EFEFEF}0.857 & 0.905            & \cellcolor[HTML]{EFEFEF}0.909 & 0.848            & \cellcolor[HTML]{EFEFEF}0.850 \\
\multirow{-4}{*}{\rotatebox{90}{DIS-VD}}  & $E_M^{\phi} \uparrow$ & 0.856            & \cellcolor[HTML]{EFEFEF}0.923 & 0.873            & \cellcolor[HTML]{EFEFEF}0.880 & 0.876            & \cellcolor[HTML]{EFEFEF}0.901 & 0.931            & \cellcolor[HTML]{EFEFEF}0.941 & 0.896            & \cellcolor[HTML]{EFEFEF}0.903 \\ \bottomrule[1.2pt]
\end{tabularx}
\end{table}

\subsection{Results}
\paragraph{Results by Generated Image Count.}
We evaluate the performance of our proposed method, MaskFactory, against two state-of-the-art baselines, DatasetDM \cite{wu2024datasetdm} and Dataset Diffusion \cite{nguyen2024dataset}, on the DIS5k dataset. The DIS5k dataset comprises five sub-datasets: DIS-VD, DIS-TE1, DIS-TE2, DIS-TE3, and DIS-TE4. We use the robust segmentation method IS-Net \cite{qin2022highly} as a baseline for the DIS task. For the two baselines, DatasetDM and Dataset Diffusion, we use the default parameters provided in their open-source implementations.

First, we identify the best-performing model on the DIS-VD validation set and then evaluate its performance on the other sub-datasets. The models are trained on the DIS-TR dataset, which is augmented with generated datasets of varying sizes: 2500, 5000, 7500, and 10000 images.

The experimental results are presented in Table \ref{tab:main1}. Our proposed method, MaskFactory, consistently outperforms the baselines across all sub-datasets and evaluation metrics. As the number of generated images increases, the performance of MaskFactory improves, achieving the best results with 10000 generated images. Notably, MaskFactory attains the highest $max F_1$ scores across all sub-datasets, with improvements ranging from 0.044 to 0.052 compared to the IS-Net baseline.

In contrast, while DatasetDM and Dataset Diffusion also show some performance gains, they encounter the issue of collapse when the generated data exceeds 5000 images. In these cases, the segmentation network's performance stagnates or even degrades. For the DIS segmentation task, the use of pseudo-labels could introduce additional errors, leading to performance declines.

Furthermore, MaskFactory demonstrates superior performance in terms of the weighted F-measure ($F_{\beta}^{\omega}$), S-measure ($S_{\alpha}$), and Enhanced-alignment Measure ($E_M^{\phi}$). The Mean Absolute Error ($M$) is also consistently lower for MaskFactory compared to the baselines, indicating more accurate segmentation results.

\paragraph{Results by Segmentation Network.}
After achieving stable improvements in IS-Net's performance, we examined the generalizability of our approach by applying the same configuration to several state-of-the-art segmentation networks on the DIS5K dataset. Each network was trained using the DIS-TR dataset augmented with 10,000 generated image pairs. The networks considered in this study include FP-DIS \cite{zhou2023dichotomous}, UDUN \cite{pei2023unite}, BiRefNet \cite{zheng2024birefnet}, and SAMHQ \cite{ke2024segment}, all implemented with their default parameters.

The experimental results, shown in Table \ref{tab:main2}, demonstrate that our proposed method consistently enhances the performance of all evaluated networks across the five sub-datasets of DIS5K. Notably, significant improvements were observed in the $max F_1$ and $E_M^{\phi}$ metrics. For instance, on the DIS-TE1 sub-dataset, our method increased the $max F_1$ score of IS-Net from 0.740 to 0.784, FP-DIS from 0.784 to 0.805, UDUN from 0.784 to 0.799, BiRefNet from 0.866 to 0.882, and SAMHQ from 0.897 to 0.905.

Additionally, the Mean Absolute Error ($M$) decreased for all networks when applying our method, indicating more accurate segmentation results. The S-measure ($S_{\alpha}$) also consistently improved across all sub-datasets and networks, highlighting the effectiveness of our approach in capturing structural similarity.

On the DIS-VD sub-dataset, used for validation, our method boosted the $max F_1$ score of IS-Net from 0.791 to 0.835, FP-DIS from 0.823 to 0.851, UDUN from 0.823 to 0.847, BiRefNet from 0.897 to 0.905, and SAMHQ from 0.842 to 0.847. These findings underscore the generalizability of our approach, as it enhances the performance of diverse segmentation networks without requiring any network-specific modifications.

\paragraph{Visual Results.}

\begin{wraptable}{r}{0.4\textwidth}
\vspace{-0.4cm}
\fontsize{8.5}{8}\selectfont
\centering
\renewcommand\tabcolsep{2.5pt}
\caption{Similarity with original dataset.}
\vspace{-0.2cm}
\label{tab:tsne}
\begin{tabular}{r|cc}
\toprule[1.2pt]
\textbf{Generate Type} & \textbf{CLIP} & \textbf{UMAP} \\ \midrule
DatasetDM \cite{wu2024datasetdm} & 0.6683 & 0.7017 \\
Dataset Diffusion \cite{nguyen2024dataset} & 0.6217 & 0.6349 \\
MaskFactory(rigid) & 0.8791 & 0.8961 \\
MaskFactory(non-rigid) & \textbf{0.9147} & \textbf{0.9346} \\
MaskFactory(All) & 0.8967 & 0.9103 \\ \bottomrule[1.2pt]
\end{tabular}
\end{wraptable}
As shown in Figure \ref{fig:visual results} (Appendix), our approach achieves precise results, comprising both rigid and non-rigid transformations. The non-rigid transformations, illustrated in columns (b) and (c), enable shape editing, such as removing a table corner or merging two backpacks into one. In contrast, the rigid transformations, demonstrated in columns (d), (e), and (f), primarily involve viewpoint changes, showcasing the original mask rotated in 3D space. Notably, our method effectively preserves the topological structure information of the original image, including the holes on the chair back. This allows for the low-cost generation of high-precision, diverse data pairs.
\begin{figure}[t]
    \centering
    \begin{subfigure}[b]{1.0\textwidth}
        \centering
        \includegraphics[width=\textwidth]{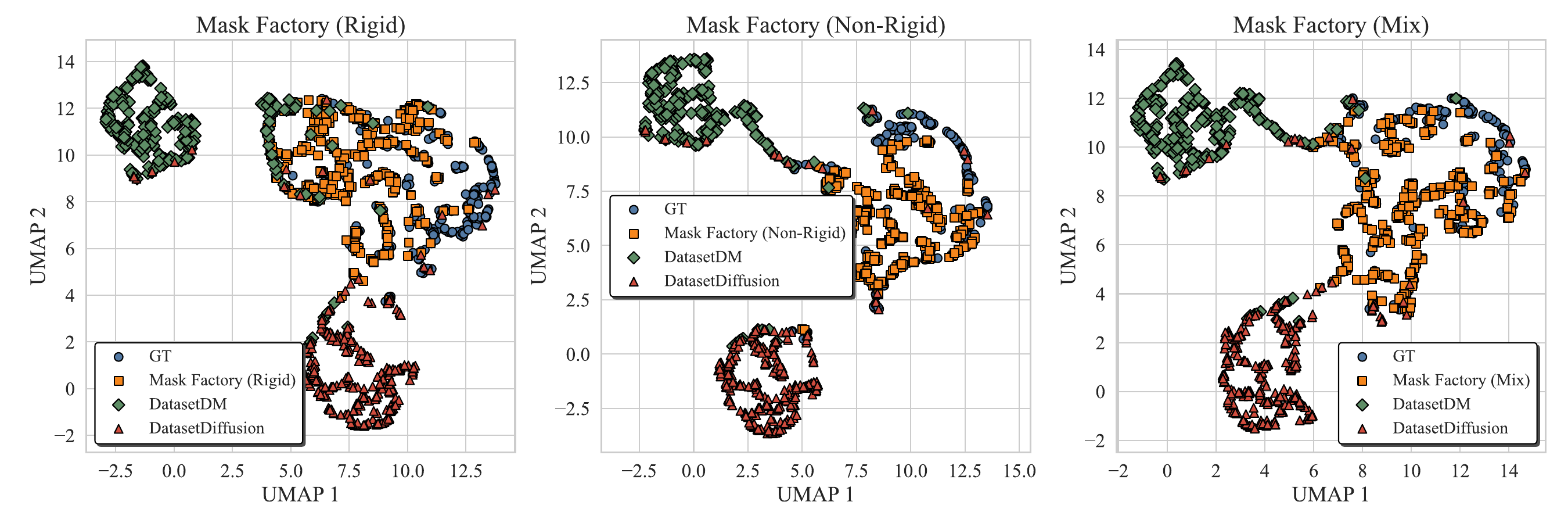}
        \vspace{-0.6cm}
        \caption{Differences in UMAP distribution of generated mask images}
        \vspace{-0.3cm}
        \label{fig:fig1}
    \end{subfigure}
    \vskip\baselineskip
    \begin{subfigure}[b]{1.0\textwidth}
        \centering
        \includegraphics[width=\textwidth]{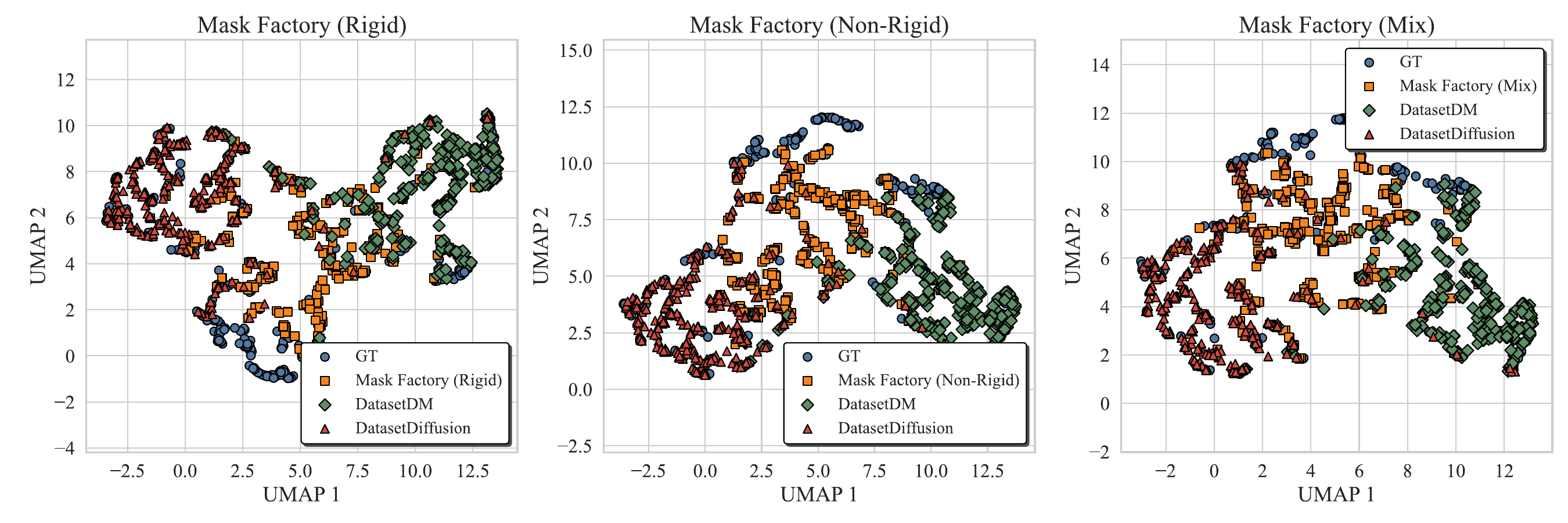}
         \vspace{-0.6cm}
        \caption{Differences in UMAP distribution of generated corresponding RGB images}
        \label{fig:fig2}
    \end{subfigure}
    \caption{UMAP Distribution Differences}
    \vspace{-0.2cm}
    \label{fig:umap_differences}
\end{figure}
To investigate differences between generated and real images, we analyzed image and mask distributions. Specifically, we used the CLIP model\cite{radford2021learning} to extract features from 300 real images, as well as images and masks generated by DatasetDM, DatasetDiffusion, and MaskFactory. We then applied UMAP\cite{mcinnes2018umap} for dimensionality reduction on these 300 features. The feature distributions of masks and images are visualized in Figures \ref{fig:umap_differences}(a) and \ref{fig:umap_differences}(b), respectively.

In the mask editing domain, MaskFactory demonstrates superior mask fidelity compared to other generation methods, primarily due to the incorporation of topological consistency constraints. This results in a feature distribution that closely aligns with that of real images. Conversely, for RGB images, the prior from diffusion VAE introduces a larger disparity between the generated and real image distributions. However, the distribution generated by MaskFactory shows a greater overlap with the real image distribution compared to the other two methods.

Furthermore, we quantified the differences using cosine similarity, as presented in Table \ref{tab:tsne}. The results indicate that our method achieves the closest distribution to real images, further validating the effectiveness of MaskFactory in generating realistic masks and images.
\vspace{-0.3cm}


\section{Discussion}

\subsection{Ablation Study}
\paragraph{Mask Generation Type Ablation.}
\begin{wraptable}{r}{0.4\textwidth}
\vspace{-0.4cm}
\fontsize{8.5}{9.6}\selectfont
\centering
\renewcommand\tabcolsep{2.5pt}
\caption{Ablations on generation types.}
\vspace{-0.2cm}
\label{tab:ablation_data_generation}
\begin{tabular}{rcccc}
\toprule[1.2pt]
\textbf{Type}      & $max F_1 \uparrow$ & $M \downarrow$   & $S_{\alpha} \uparrow$ & $E_M^{\phi} \uparrow$ \\ \midrule
\textbf{Rigid}     & 0.768    & 0.074 & 0.807   & 0.867                     \\
\textbf{Non-Rigid} & 0.771    & 0.074 & 0.796   & 0.858                     \\
\rowcolor[HTML]{EFEFEF} 
\textbf{Mix}       & 0.784    & 0.073 & 0.829   & 0.875                     \\ \bottomrule[1.2pt]
\end{tabular}
\vspace{-0.3cm}
\end{wraptable}
In our study, we implemented mask rigid editing, non-rigid editing, and mixed editing—each leveraging our novel mask control technology tailored for specific application scenarios. Rigid editing is designed for scenarios requiring precise geometric adjustments, primarily focusing on viewpoint and scale transformations. Non-rigid editing caters to applications needing high adaptability, handling topologically consistent deformations and complex, dynamic image edits. Mixed editing combines the advantages of both approaches, offering a comprehensive solution. We further evaluated the performance gains of each editing strategy. Our experimental results, as shown in Table \ref{tab:ablation_data_generation}.

From the table, we observe that mixed editing achieves the highest performance across most metrics. Specifically, it achieves the highest $max F_1$ score and $S_{\alpha}$, indicating superior structural fidelity and segmentation quality. The slight improvement in $M$ and $E_M^{\phi}$ further underscores the versatility and effectiveness of mixed editing in creating diverse and realistic image-mask combinations.

\paragraph{Loss Function Ablation.}
We introduce content and structure losses into our model. The Discriminative Loss, implemented via a discriminator, evaluates the differences between generated and real images, aiming to enhance the realism and quality of the outputs. The Edge Constraint Loss focuses on maintaining edge coherence during image editing, which is critical for preserving detailed structural information. We conducted ablation experiments to evaluate the impact of each loss function. The experimental results, shown in Table \ref{tab:ablation_loss_functions}.
\begin{wraptable}{r}{0.4\textwidth}
\vspace{-0.2cm}
\fontsize{8.5}{9.6}\selectfont
\centering
\renewcommand\tabcolsep{2.5pt}
\caption{Ablations on loss functions.}
\vspace{-0.2cm}
\label{tab:ablation_loss_functions}
\begin{tabular}{ccc|cc}
\toprule[1.2pt]
\multicolumn{3}{c|}{Loss Function} & \multirow{2}{*}{$max F_1 \uparrow$} & \multirow{2}{*}{$M \downarrow$} \\
$\mathcal{L}_{\text{GAN}}$ & $\mathcal{L}_{\text{content}}$ & $\mathcal{L}_{\text{structure}}$ &                           &                      \\ \midrule
$\checkmark$  &            &              & 0.778                     & 0.073                \\
       & $\checkmark$      &              & 0.745                     & 0.075                \\
       &            & $\checkmark$        & 0.751                     & 0.074                \\
       \rowcolor[HTML]{EFEFEF} 
$\checkmark$  & $\checkmark$      & $\checkmark$        & 0.784                     & 0.073                \\ \bottomrule[1.2pt]
\end{tabular}
\vspace{-0.6cm}
\end{wraptable}
From the table, we observe that the combination of all three loss functions ($\mathcal{L}_{\text{GAN}}$, $\mathcal{L}_{\text{content}}$, and $\mathcal{L}_{\text{structure}}$) achieves the highest $max F_1$ score, indicating the best performance in terms of structural fidelity and realism.

\begin{wraptable}{H}{0.4\textwidth}
\vspace{-0.4cm}
\fontsize{8.5}{9.6}\selectfont
\centering
\renewcommand\tabcolsep{2.5pt}
\caption{Ablations on conditions.}
\vspace{-0.2cm}
\label{tab:ablation_generated}
\begin{tabular}{ccc|cc}
\toprule[1.2pt]
Mask & Prompt & Canny & $max F_1 \uparrow$ & $M \downarrow$   \\ \midrule
$\checkmark$ &        &       & 0.778    & 0.075 \\
$\checkmark$ & $\checkmark$   &       & 0.782    & 0.073 \\
$\checkmark$ &        & $\checkmark$  & 0.764    & 0.080 \\
\rowcolor[HTML]{EFEFEF}
$\checkmark$ & $\checkmark$   & $\checkmark$  & 0.784    & 0.073 \\ \bottomrule[1.2pt]
\end{tabular}
\end{wraptable}

\subsection{Limitation}
\label{sec:limit}
Despite the favorable outcomes achieved by our method, it still encounters significant issues. Although we experimented with different conditions, the results are shown in Table \ref{tab:ablation_generated}. ControlNet sometimes produces unnatural images with stark foreground-background distinctions, necessitating additional harmonization. Complex scenarios can yield unrealistic elements, such as improperly positioned objects. Additionally, our method relies on pre-annotated image-mask pairs, limiting its ability to generate data autonomously and requiring high-quality initial annotations.
\subsection{Conclusion}
This paper introduces MaskFactory, a novel two-stage approach for generating high-quality synthetic datasets for DIS tasks. By combining rigid and non-rigid mask editing techniques and using multi-conditional control for image generation, MaskFactory produces diverse and precise synthetic image-mask pairs, significantly reducing dataset preparation time and costs. Experiments on the DIS5K benchmark demonstrate the superior performance of MaskFactory compared to existing methods in terms of quality and efficiency.

\clearpage
\section*{Acknowledgements}
We thank Yuyan Huang, Hong Liu, and Wenjun Ji for fruitful discussions during the course of this project. Haoqian Qian and Xiaogang Jin were supported by Key R\&D Program of Zhejiang (No.2024C01069). Deng-Ping Fan was supported by NSFC (No.62476143).

\bibliographystyle{unsrt} 

\clearpage  
\appendix
\begin{center}
    \Large \textbf{Appendix}
\end{center}
\section{Pseudocode for the \ourmodel{} Algorithm}
\ourmodel{} is a two-stage approach for generating high-quality synthetic datasets for DIS tasks. In the first stage, existing ground truth masks undergo rigid and non-rigid editing to generate diverse synthetic masks. Rigid editing uses geometric priors from diffusion models for precise viewpoint transformations, while non-rigid editing employs adversarial training and self-attention mechanisms for complex shape modifications while preserving topology.

In the second stage, the generated masks and their corresponding Canny edges serve as conditions, along with category prompts, to guide the generation of high-resolution RGB images using a multi-conditional control generation method. This process ensures consistency between the generated images and masks while enhancing dataset realism and diversity. Our pseudocode visible Algorithm \ref{alg_mask}

\begin{algorithm}[H]
\caption{\ourmodel{} Algorithm}
\label{alg_mask}
\SetKwInOut{Input}{Input}
\SetKwInOut{Output}{Output}

\Input{$D = \{(I^r_i, I^m_i)\}_{i=1}^N$ - Original dataset}
\Output{$G = \{(g^r_i, g^m_i)\}_{i=1}^M$ - Synthetic dataset}

\BlankLine
\textbf{Stage 1: Mask Editing} \\
\For{$i \gets 1$ \KwTo $N$}{
    \textbf{Step 1.1: Rigid Mask Editing} \\
    $I^{m'}_i \gets \text{Invert}(I^m_i)$ \tcp*[h]{Invert source mask} \\
    $g^m_i \gets \psi_\theta(I^{m'}_i, T_i)$ \tcp*[h]{Apply viewpoint transformation}
    
    \BlankLine
    \textbf{Step 1.2: Non-Rigid Mask Editing} \\
    $E_s \gets E(I^m_i)$ \tcp*[h]{Extract edge map from source mask} \\
    $\mathcal{V} \gets \{v_j\}^{N_v}_{j=1}$ \tcp*[h]{Obtain key points from edge map} \\
    $\mathcal{T}_s \gets (V, E_s)$ \tcp*[h]{Construct source mask structural graph} \\
    $g^m_i \gets G_\theta(z, P_i, I^m_i)$ \tcp*[h]{Generate synthetic mask} \\
    ${T}_g = (\mathcal{V}, \mathcal{E}_s)$ \tcp*[h]{Construct synthetic mask structural graph} \\
    
    \BlankLine
    $\mathcal{L}_{GAN} \gets \mathbb{E}_{T_s \sim p_{data}(T_s)}[\log D_\phi(T_s)] + \mathbb{E}_{T_g \sim p_{gen}(T_g)}[\log(1 - D_\phi(T_g))]$ \\
    $\mathcal{L}_{content} \gets \|g^m_i - I^m_i\|_1$ \\
    $\mathcal{L}_{structure} \gets \|T_g - T_s\|_1$ \\
    $\mathcal{L}_{total} \gets \mathcal{L}_{GAN} + \lambda_1 \mathcal{L}_{content} + \lambda_2 \mathcal{L}_{structure}$ \\
    Update $G_\theta$ and $D_\phi$ to minimize $\mathcal{L}_{total}$
}

\BlankLine
\textbf{Stage 2: Image Generation} \\
\For{$i \gets 1$ \KwTo $M$}{
    $c^s_i \gets g^m_i$ \tcp*[h]{Segmentation condition} \\
    $c^y_i \gets \text{Canny}(g^m_i)$ \tcp*[h]{Canny condition} \\
    $z \sim \mathcal{N}(0, 1)$ \tcp*[h]{Sample Gaussian noise} \\
    $g^r_i \gets M_\theta(z, B_\theta(c^s_i), B_\theta(c^y_i))$ \tcp*[h]{Generate RGB image}
}

\BlankLine
\Return $G = \{(g^r_i, g^m_i)\}_{i=1}^M$
\end{algorithm}

\section{Dataset Details}
\label{datasetDetails}
We conduct our experiments on the DIS5K dataset, which comprises 5,479 high-resolution images featuring camouflaged, salient, or meticulous objects in various backgrounds. The DIS5K dataset is divided into three subsets: DIS-TR (3,000 images) for training, DIS-VD (470 images) for validation, and DIS-TE (2,000 images) for testing. For data augmentation, we utilize the mask portion of the training subset (DIS-TR).

To evaluate our models, we employ a diverse set of metrics to ensure comprehensive performance assessment. These metrics include {max F\(_1\)}, which balances precision and recall, providing a harmonic mean that is indicative of overall accuracy; {F\(_\beta^\omega\)}, a weighted F-measure that compensates for class imbalances, with values ranging from 0 to 1, where higher values denote superior performance; {$M$} (Mean Absolute Error), which calculates the average absolute difference between the predicted and ground truth masks, with lower values signifying better accuracy; {S\(_\alpha\)}, a structural similarity measure that evaluates the preservation of significant structures within the image, with values closer to 1 indicating better performance; and {E\(_M^\phi\)}, an enhanced measure that considers both pixel-level and image-level information for a more holistic evaluation, where higher values represent better performance. Collectively, these metrics provide a robust framework for assessing the effectiveness and reliability of our segmentation models.

\section{Mathematical Details}
\label{math}
\subsection{Diffusion-Based Image Generation}

The diffusion model employed in {\ourmodel{}} for image generation follows the formulation introduced by Ho et al.~\cite{ho2020denoising}. Given a data distribution $x_0 \sim q(x_0)$, the forward diffusion process is defined as a Markov chain that gradually adds Gaussian noise to the data:

\begin{equation}
q(x_t|x_{t-1}) = \mathcal{N}(x_t; \sqrt{1-\beta_t}x_{t-1}, \beta_t\mathbf{I}),
\end{equation}

where $\beta_t \in (0, 1)$ is a variance schedule. The reverse process is learned by a neural network $\epsilon_\theta$ that predicts the noise added at each step:

\begin{equation}
p_\theta(x_{t-1}|x_t) = \mathcal{N}(x_{t-1}; \mu_\theta(x_t, t), \sigma_t^2\mathbf{I}),
\end{equation}

where $\mu_\theta(x_t, t) = \frac{1}{\sqrt{\alpha_t}} \left(x_t - \frac{\beta_t}{\sqrt{1-\bar{\alpha}_t}} \epsilon_\theta(x_t, t)\right)$ and $\alpha_t = 1 - \beta_t$, $\bar{\alpha}_t = \prod_{s=1}^t \alpha_s$.

The objective is to maximize the \textbf{variational lower bound}:

\begin{equation}
\mathcal{L} = \mathbb{E}_{q(x_0)} \mathbb{E}_{q(x_1,...,x_T|x_0)} \left[\sum_{t=1}^T \log \frac{p_\theta(x_{t-1}|x_t)}{q(x_{t-1}|x_t,x_0)}\right].
\end{equation}

\subsection{Topology-Preserving Adversarial Training}

The \textbf{topology-preserving adversarial training} in \textbf{\ourmodel{}} involves a generator $G_\theta$ and a discriminator $D_\phi$. The generator aims to minimize the adversarial loss:

\begin{equation}
\mathcal{L}_{\text{GAN}}(G) = \mathbb{E}_{T_g \sim p_{\text{gen}}(T_g)}[\log(1 - D_\phi(T_g))],
\end{equation}

while the discriminator tries to maximize the adversarial loss:

\begin{equation}
\mathcal{L}_{\text{GAN}}(D) = \mathbb{E}_{T_s \sim p_{\text{data}}(T_s)}[\log D_\phi(T_s)] + \mathbb{E}_{T_g \sim p_{\text{gen}}(T_g)}[\log(1 - D_\phi(T_g))].
\end{equation}

The \textbf{generator} and \textbf{discriminator} are updated alternately to reach an equilibrium.

\subsection{Structural Graph Construction}

To preserve the topological structure of the source masks during editing, \textbf{\ourmodel{}} constructs structural graphs $T_s$ and $T_g$ for the source and synthetic masks, respectively. The structural graph $T = (V, E)$ consists of a set of vertices $V = \{v_j\}_{j=1}^{N_v}$ representing key points and a set of edges $E$ representing their connectivity.

The structure preservation loss is defined as the {L1 distance} between the structural graphs:

\begin{equation}
\mathcal{L}_{\text{structure}} = \|T_g - T_s\|_1 = \sum_{(i,j) \in E} \|T_g(i,j) - T_s(i,j)\|,
\end{equation}

where $T_g(i,j)$ and $T_s(i,j)$ denote the edge weights between vertices $i$ and $j`$ in the synthetic and source mask structural graphs, respectively.

By minimizing the \textbf{structure preservation loss} along with the adversarial and content losses, \textbf{\ourmodel{}} ensures that the edited masks maintain the topological structure of the source masks.

\section{Visualization of the results using two editing methods from \ourmodel{}.}
In this section, we demonstrate \ourmodel{}'s ability to edit masks for common and fine-grained objects.

\subsection{Visualization of Common Object Mask Editing}

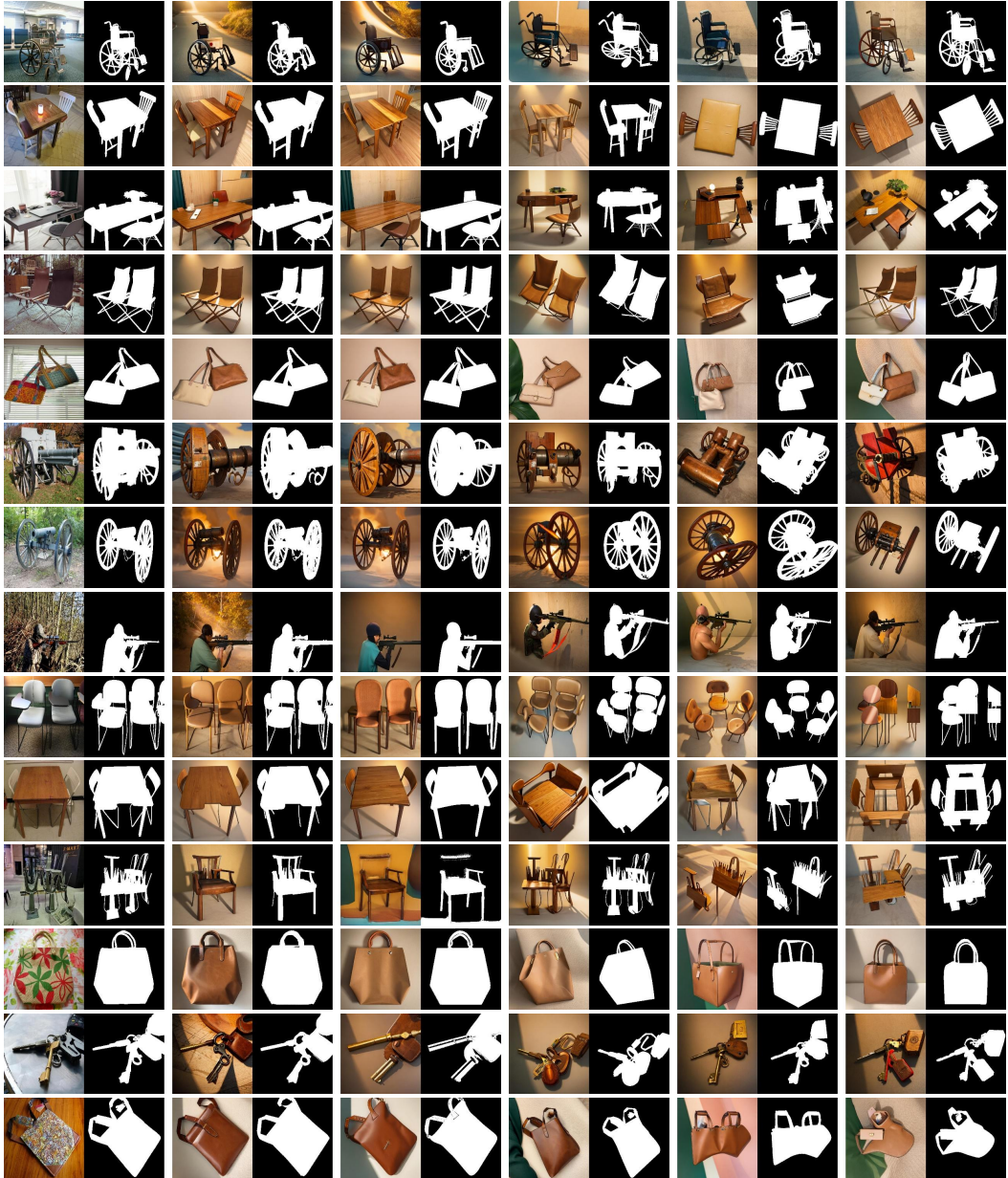
\begin{figure}[t]
    \centering
    \includegraphics[width = \linewidth]{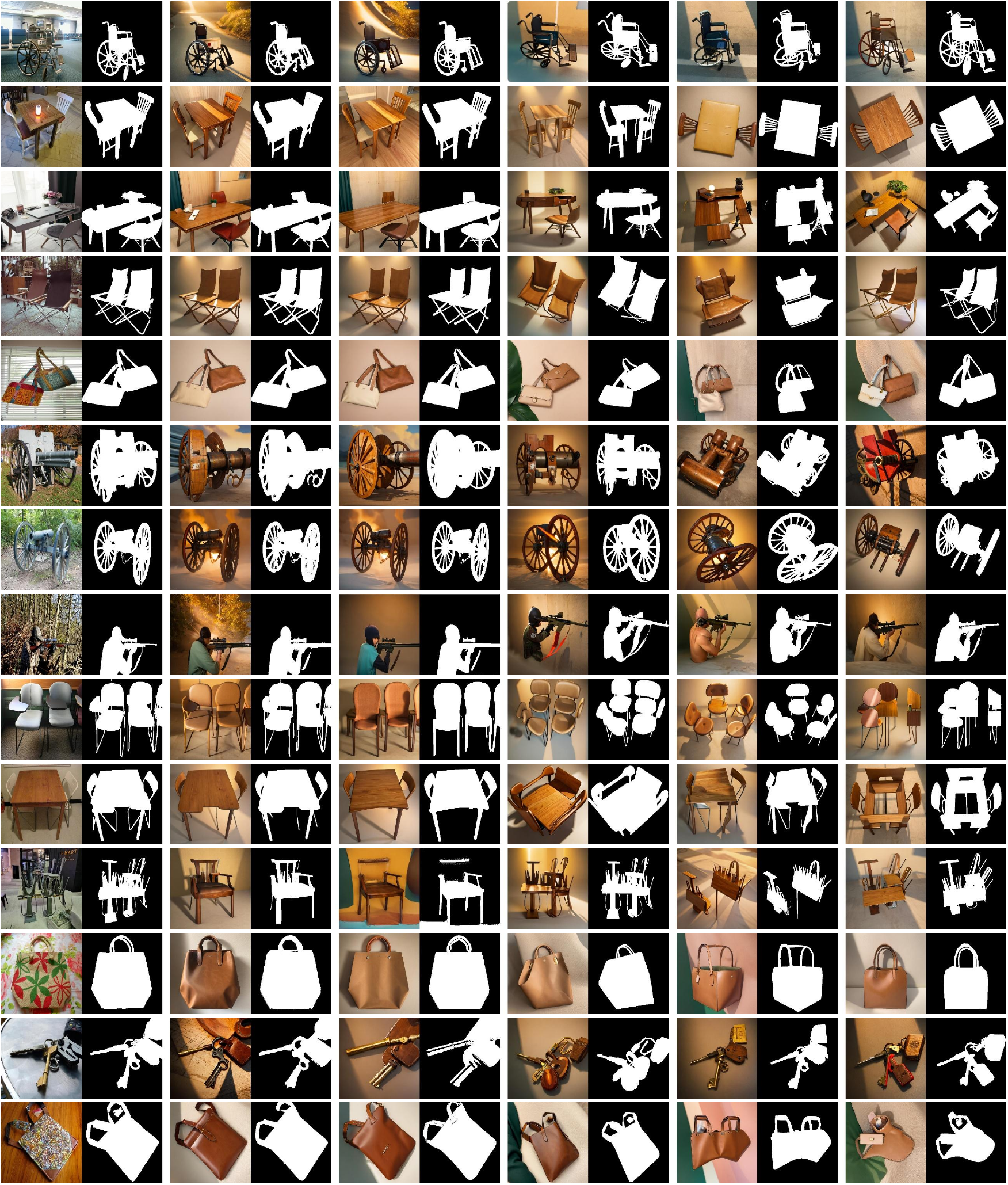}
    \put(-0.99\linewidth, -1em){(a) raw image}
    \put(-0.80\linewidth, -1em){(b) rigid 1}
    \put(-0.63\linewidth, -1em){(c) rigid 2}
    \put(-0.5\linewidth, -1em){(d) non-rigid 1}
    \put(-0.32\linewidth, -1em){(e) non-rigid 2}
    \put(-0.16\linewidth, -1em){(f) non-rigid 3}
    \caption{Visual results of common object mask editing. The model demonstrates strong topological structure preservation and diverse editing outcomes with both rigid and non-rigid edits.}
    \label{fig:visual results}
\end{figure}

We selected common household items such as tables, chairs, bags, and musical instruments for editing using \ourmodel{}. Non-rigid edits were performed with different prompts, and rigid edits were performed from different viewpoints. The editing results and corresponding RGB image generations are shown in Figure~\ref{fig:visual results}. \ourmodel{} exhibits strong topological structure preservation and produces diverse editing outcomes, as demonstrated by the variety of modifications made to both rigid and non-rigid objects.

\subsection{Visualization of Fine-Grained Object Mask Editing}

\begin{figure}[t]
    \centering
    \includegraphics[width=\linewidth]{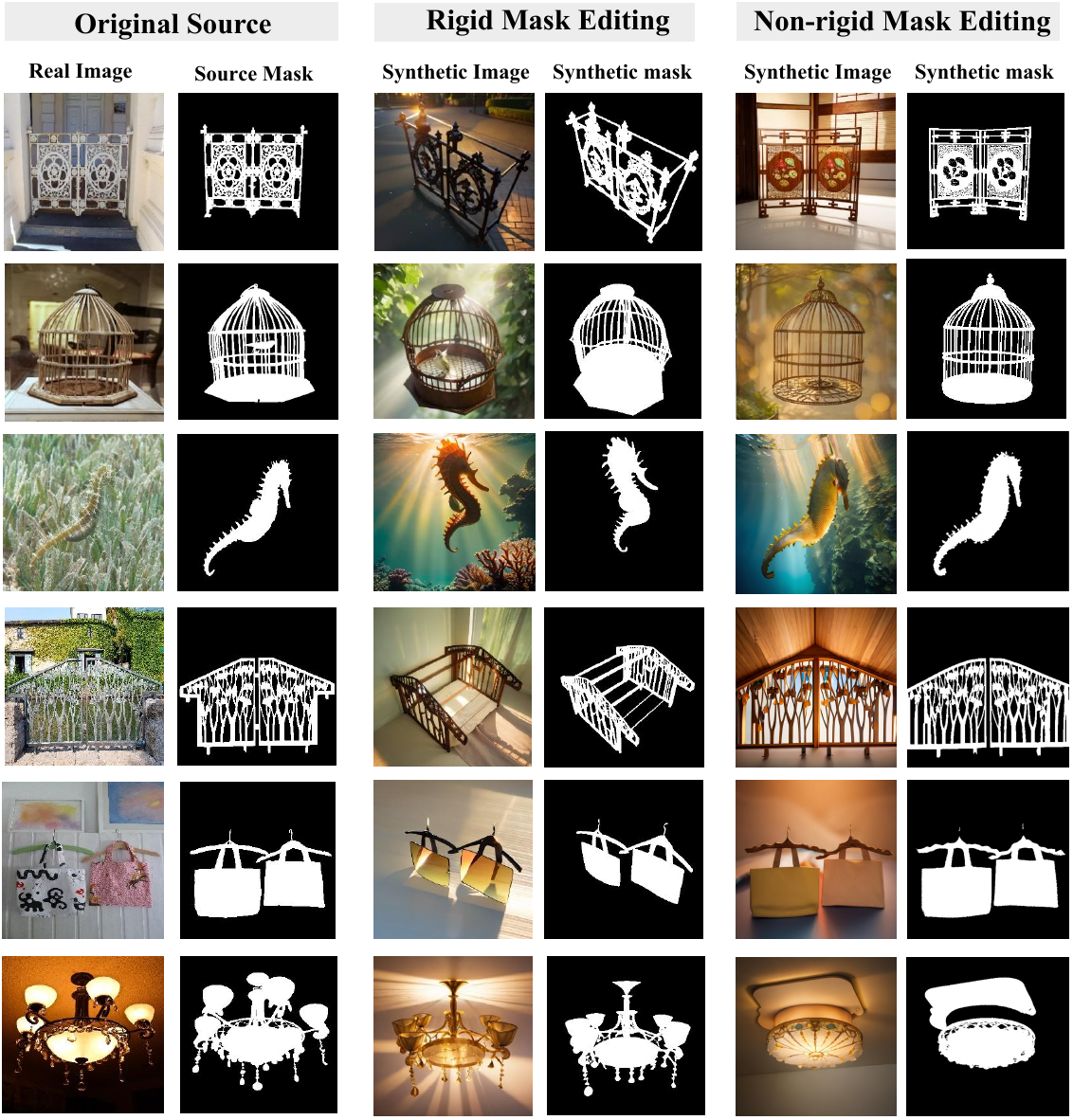}
    \caption{Visual results of fine-grained object mask editing. The model successfully edits complex structures without compromising the original mask's semantic information.}
    \label{fig:fine0523}
\end{figure}

We selected fine-grained objects from \ourmodel{}'s generated masks, such as ornate European chandeliers, iron gates, birdcages, and seahorses, to showcase \ourmodel{}'s detailed editing capabilities. Even with complex geometries, \ourmodel{} can perform topology-preserving edits without losing the original mask's semantic information. These intricate structures play a crucial role in segmentation metrics. As illustrated in Figure~\ref{fig:fine0523}, the visual results demonstrate the model's ability to handle fine-grained details while maintaining the integrity of the original mask's structure.

\subsection{Visualization Results with Canny Constraints}

\begin{figure}[t]
    \centering
    \includegraphics[width=0.8\linewidth]{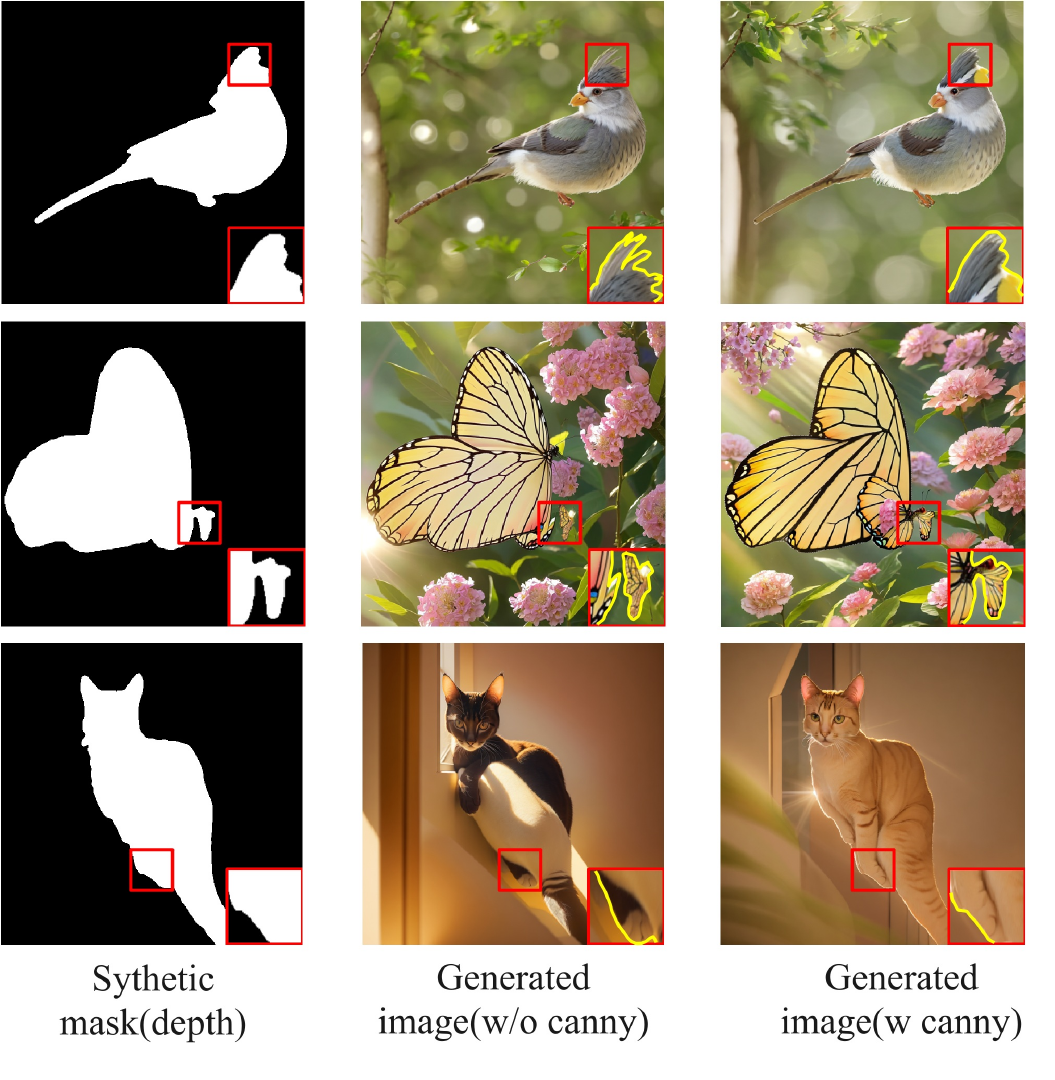}
    \caption{Canny condition visual results. The generated images show improved boundary precision and better structural coherence.}
    \label{fig:fine051123}
\end{figure}

After incorporating Canny edge detection as a constraint, the visual results of MaskFactory show a significant improvement in boundary precision. The Canny edges effectively guide the generation process, resulting in images with clearer and more accurate boundary details, avoiding vague or ambiguous transition areas. The visualizations demonstrate that the Canny edges not only better constrain the contours of the generated images but also enhance the overall fidelity and visual quality of the output. Compared to models without edge constraints, our approach produces more detailed and structurally coherent images, as shown in Figure~\ref{fig:fine051123}.

\subsection{Topological Structure Visualization}

\begin{figure}[t]
    \centering
    \includegraphics[width=\linewidth]{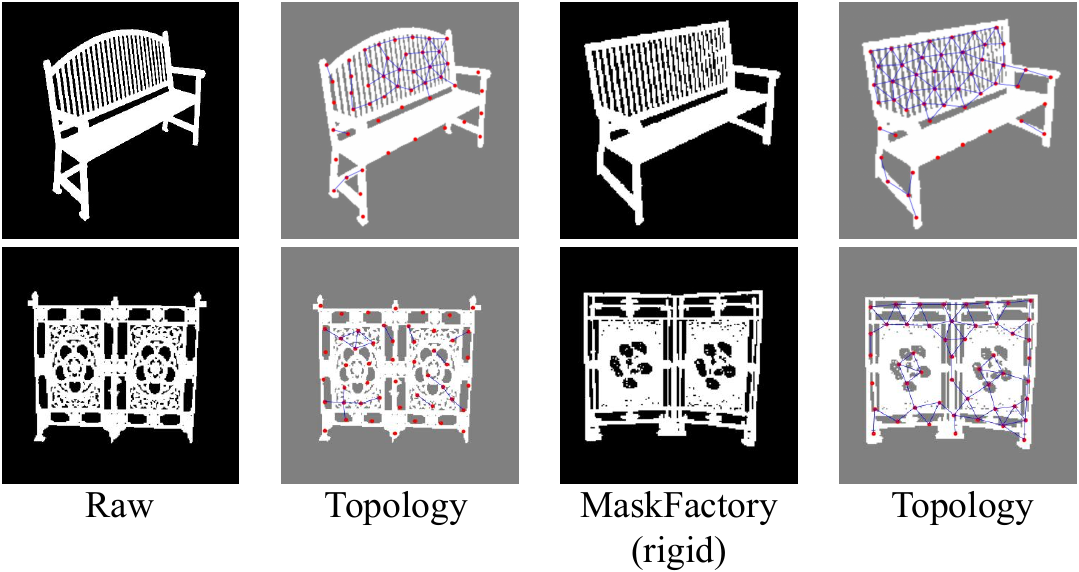}
    \caption{Topological Structure Visualization. The visualizations demonstrate the model's ability to maintain and manipulate topology during the editing process.}
    \label{fig:fine05213}
\end{figure}

We performed topological structure visualizations on selected samples to assess the model's ability to preserve topology during the editing process. These visualizations clearly show how our method retains the topological structure of the original data while allowing for effective manipulation when necessary. Whether dealing with complex geometric shapes or subtle structural modifications, the topological visualizations illustrate that our approach reliably preserves geometric consistency and topological features throughout the editing process, as demonstrated in Figure~\ref{fig:fine05213}.

\section{Future Work}
Currently, our work has merely explored the performance of the generation method, MaskFactory, within the DIS segmentation task, which can be regarded as a form of data augmentation. However, this method actually has the potential to be applied more extensively to a wider variety of tasks, such as pose estimation \cite{shen2024imagpose,shenadvancing,li2024deviation,li2024translating}, ultra-high-resolution image segmentation \cite{sun2024ultrahighresolutionsegmentationboundaryenhanced,sunprogram,yin2024class}, remote sensing image enhancement \cite{ma2024logcanadaptivelocalglobalclassaware,10095835,tao2023dudb}, image topological enhancement \cite{RAMMVC,scMFC}, image super-resolution \cite{zhang2024cf,zhang2023multi}, autonomous driving \cite{Zhang2023EHSSAE,zhang2024mapexpertonlinehdmap}, and multi-view estimation \cite{Yuan2024h,Yuan2024i,Yuan2024j}. Moreover, generation-based methods will inevitably introduce some noisy data. Thus, how to screen the generated data or study how to learn from the noisy data and evaluate the quality of the generated data is also a point worthy of attention.


\end{document}